\documentclass{article}

\PassOptionsToPackage{numbers, compress}{natbib}

\usepackage[final]{neurips_2019}

\usepackage[utf8]{inputenc} 
\usepackage[T1]{fontenc}    
\usepackage{hyperref}       
\usepackage{url}            
\usepackage{booktabs}       
\usepackage{amsfonts}       
\usepackage{nicefrac}       
\usepackage{microtype}      
\usepackage{graphicx} 
\usepackage{adjustbox}
\usepackage{tabularx}
\usepackage{subcaption} 
\usepackage{float}

\title{Investigation of Error Simulation Techniques for Learning Dialog Policies for Conversational Error Recovery}

\author{
Maryam Fazel-Zarandi\thanks{Equal contributions.}\\
Amazon Inc, USA\\
\texttt{fazelzar@amazon.com}\\
\And
Longshaokan Wang\textsuperscript{*} \\
Amazon Inc, USA \\
\texttt{longsha@amazon.com} \\
\AND
Aditya Tiwari\\
Amazon Inc, USA \\
\texttt{aditiwar@amazon.com}
\And
Spyros  Matsoukas\\
Amazon Inc, USA \\
\texttt{matsouka@amazon.com}}

\begin{document}

\maketitle

\begin{abstract}
Training dialog policies for speech-based virtual assistants requires a plethora of conversational data. 
The data collection phase is often expensive and time consuming due to human involvement. 
To address this issue, a common solution is to build user simulators for data generation. 
For the successful deployment of the trained policies into real world domains, it is vital that the user simulator mimics realistic conditions. 
In particular, speech-based assistants are heavily affected by automatic speech recognition and language understanding errors, hence the user simulator should be able to simulate similar errors. 
In this paper, we review the existing error simulation methods that induce errors at audio, phoneme, text, or semantic level; 
and conduct detailed comparisons between the audio-level and text-level methods. 
In the process, we improve the existing text-level method by introducing confidence score prediction and out-of-vocabulary word mapping. 
We also explore the impact of audio-level and text-level methods on learning a simple clarification dialog policy to recover from errors to provide insight on future improvement for both approaches.
\end{abstract}

\section{Introduction}

Spoken dialog agents are becoming a prevalent tool in personal assistant devices. 
Such human-like interfaces create a rich experience for users by enabling them to complete many tasks hands-free and eyes-free in a conversational manner. 
Depending on how complicated a task is, multiple rounds of conversation may be needed for the assistant to fully understand user requests. 
To achieve this, these interfaces require the design of complex dialog policies which can generate appropriate responses to queries and steer the conversation. 
Designing dialog policies is difficult and requires a plethora of data, because responding to a user is conditioned on the task, 
the user goals and preferences, and the dialog history. Additionally, environmental noise and ambiguous user utterances add to the complexity. 
Furthermore, collecting data is expensive and time consuming due to human involvement. 
It may also be difficult to get engagement from real users in the early stages of the dialog policy development. 
Consequently, it is necessary to leverage semi-supervised or unsupervised methods 
which require less human labeling and can account for many branching user interactions. 
Reinforcement learning (RL) has become one such method for training dialog agents 
(e.g., \cite{Singh2002, Williams2007, Lee2012}).
However, RL training may require the use of a user simulator which can account for realistic branching user paths \cite{Schatzmann2006}. 

Ensuring a successful transfer of an RL policy from simulation to production requires a moderately accurate simulation \cite{Tan2018, Sadeghi2017, Chebotar2018}. 
To improve the realism of RL policies, current methods in user simulators induce errors at different levels to mimic noisy communication channels. 
More specifically, existing methods induce errors at audio \cite{Fazel2017}, phoneme \cite{Stuttle2004, Wang2018}, 
text \cite{Pietquin2005, Schatzmann2007, Simonnet2018} or semantic \cite{Quarteroni2010, Thomson2012, XiujunLi2017} level. 
In this paper, we compare different error simulation methods 
and investigate which method simulates the most realistic errors and is thus most suitable for RL-like tasks; 
we also improve the existing methods in the process.
The audio and text level approaches are the easiest to operationalize and generalize to new domains (see Section \ref{sec:background}), 
and hence will be our focus.

The contributions of this paper are as follows: 
First, we conduct a comparative study among different error simulation methods. 
In particular, we evaluate audio- and text-level methods in terms of 
word error distribution, automatic speech recognition (ASR) confidence score distribution, whether the produced errors are distinguishable from real errors, and their effect on natural language understanding. 
We show that the text-level method is comparable to the audio-level method in the above metrics, 
and should be the preferred method for its speed and simplicity.
Second, we improve the existing text-level method by introducing ASR confidence score prediction and out-of-vocabulary word mapping.
Confidence scores are important signals for error recovery policies.
To this end, we present a simple method 
to predict ASR confidence scores from the simulated ASR output. 
To our knowledge 
we are the first to tackle the problem of score prediction from the simulated ASR output in the absence of audio signals. 
Third, we improve the discriminator approach used by \cite{Wang2018} for evaluating error simulation, 
by including predicted ASR score and word error rate (WER) features as part of the input.
Lastly, we explore the impact of audio- and text-level methods on learning a simple clarification policy by evaluating them on held-out annotated data and conducting a user study.

\section{Background}
\label{sec:background}

When a user talks with a conversational agent, 
the audio signal is first translated into text using an ASR module. 
Using a natural language understanding (NLU) module, the agent extracts meaning from a possibly noisy input, 
usually in terms of user intents and slots, where an intent indicates the user's intention (e.g., getting the plot) and slots represent information about a particular entity (e.g., movie title). 
Finally, the agent decides what to respond to the user based on its dialog policy. 
To be successful, the dialog policy should allow the agent to recover from conversational errors such as ASR and NLU mis-recognition.

A dialog system can be formalized as a Markov Decision Process (MDP) \cite{Levin2000}. 
An MDP is a tuple <$\mathcal{S}, \mathcal{A}, \mathcal{P}, \mathcal{R}, \mathcal{\gamma}$>, where $\mathcal{S}$ is the state space, $\mathcal{A}$ is the action space, $\mathcal{P}$ is the transition probability function, $\mathcal{R}$ is the reward function, and $\mathcal{\gamma}$ is the discount factor. 
In the RL framework, at each time step \textit{t}, 
the agent observes state $s_t \in \mathcal{S}$ and selects action $a_t \in \mathcal{A}$ according to its policy ($\pi: \mathcal{S} \rightarrow \mathcal{A}$). 
After performing the selected action, the agent receives the next state \textit{$s_{t+1}$} and a scalar reward \textit{$r_t$}. 
The trajectory restarts after the agent reaches a terminal state. 
RL solvers have been used to find the optimal dialog policy (e.g., \cite{Singh2002, Williams2007, Lee2012}). 
In this context, at each turn the agent acts based on its understanding of what the user said, 
and reward function is modeled in terms of various dimensions such as per-interaction user satisfaction, accomplishment of the task, efficiency of interaction, and dialog duration. Recently, deep RL has also been applied to the problem of dialog management and has shown improvements over rule-based systems (e.g., \cite{Cuayahuitl2016, Zhao2016, Fatemi2016}).

An important challenge in using RL for learning dialog policies is creating realistic user simulators 
that can generate natural conversations similar to a human user \cite{Schatzmann2006}. 
In previous works, researchers have focused on the development of different types of user simulators (e.g., \cite{Eckert1997, Scheffler2002, Cuayahuitl2005, Georgila2006, Schatzmann2006, ElAsri2016, XiujunLi2016}). 
However, existing systems are limited with respect to accurately simulating errors that a dialog agent might face \cite{XiujunLi2017b, Fazel2017, Wang2018}. 

Errors can be simulated at audio, phoneme, text or semantic level. 
At audio level \cite{Fazel2017}, audio signals are synthesized from reference texts, noise is injected into the audio signals, 
and then noisy texts are obtained from decoding the noisy audio signals. 
This approach does not require training data, 
can leverage the text-to-speech and ASR decoding modules readily available in dialog systems, 
and can directly provide ASR confidence scores as part of the output. 
However, it is computationally expensive, and mimicking real life background noise can be difficult due to its diversity. 
Additionally, the ASR WER on synthesized speech is typically lower than on real speech. 
At phoneme level \cite{Stuttle2004, Wang2018}, reference texts are converted to phonemes, 
phoneme replacements are sampled from a phoneme confusion matrix, and then noisy phonemes are converted to noisy texts. 
This approach relies on high quality conversion modules between texts and phonemes, 
requires predicting the ASR confidence scores, 
and if a generic phoneme confusion matrix is used, the simulated errors might differ from those in a particular domain. 
At text level \cite{Pietquin2005, Schatzmann2007}, n-gram confusion matrix is constructed from transcribed training data, 
reference texts are probabilistically partitioned into n-grams, then n-gram replacements are sampled from the confusion matrix. 
This approach can directly match the distribution of error types in its output with the one in the training data, 
does not require additional modules, and is computationally efficient, but requires transcribed training data, predicting the ASR confidence scores, 
and handling of out-of-vocabulary words (we later present approaches to overcome the last two shortcomings). 
At semantic level, replacements of intents and slots are created by either sampling from a semantic confusion matrix \cite{Quarteroni2010}, 
running multiple classifiers \cite{Thomson2012}, or sampling uniformly \cite{XiujunLi2017}. 
This approach skips any conversion or decoding and is thus the fastest approach, 
but it often requires unrealistic simplifying assumptions such as the independence between intent and slot \cite{Thomson2012}, 
requires training data with semantic annotations, and is difficult to generalize to new domains.

Considering generalizability and the ease of operationalization, we focus on audio- and text-level approaches in our comparative study. 
We provide more comprehensive comparisons beyond the previous comparative study of \cite{Wang2018}, which only considers the fluency of generated alternatives and the distributional overlap between the simulated and real output on the training set.

\section{Error Simulation}

\begin{figure}[t]
  \centering
  \includegraphics[width=0.6\textwidth]{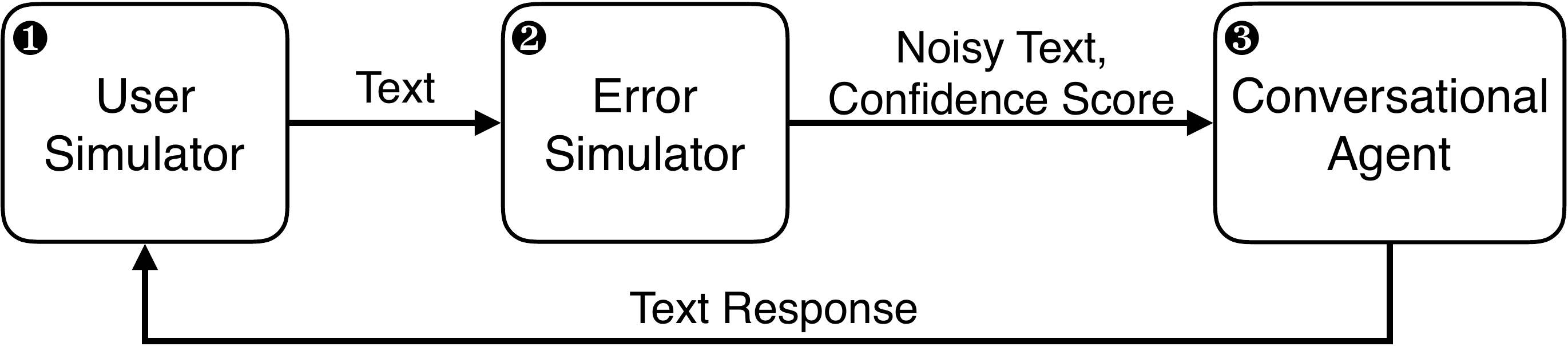}
  \caption{User simulator and conversational agent interaction. We use the text response generated by the agent before sending it to TTS to eliminate the need for ASR/NLU on the user simulator side.}
  \label{simulator-diagram}
\end{figure}

Figure \ref{simulator-diagram} illustrates the interaction between a user simulator and a conversational agent. 
The user simulator learns to behave like a human to converse with the agent 
based on the dialog context and generates noisy text using an error simulator. 
In this section we describe the audio-level and text-level error simulation techniques that we compared in this study. For text-level error simulation, we discuss our improvements to the existing approach.
  
\subsection{Audio-Level Error Simulation}

In audio-level error simulation, we take the approach of generating a spoken audio signal from text, 
adding noise into the audio and then passing the audio through an ASR model to get a hypothesis and confidence score. 
This process simulates the real-world sources of error such as background noise in a person's home. 
Concretely, we synthesize an audio signal with the text-to-speech (TTS) service provided by Amazon Polly\footnote{\url{https://aws.amazon.com/polly}} 
and contaminate the clean TTS output to achieve a desired signal-to-noise ratio (SNR). 
Contamination is performed by injecting real noise from randomly selected conversations annotated as background noise to the TTS output. 
For the purposes of this study, we use different TTS voices and dialects provided by Amazon Polly 
and select the SNR value such that we match the ASR WER of our corpus.
Finally, we pass the noisy audio through an ASR decoder to get the corresponding text and associated confidence score.

\subsection{Text-Level Error Simulation}  
\label{subsec:text-sim}

In text-level error simulation, we start with \citeauthor{Schatzmann2007}'s \cite{Schatzmann2007} n-gram confusion matrix approach (see Appendix \ref{schatzmann-approach} for details), 
which showed marginally better performance than the other error simulation approaches in \citeauthor{Wang2018}'s (\citeyear{Wang2018}) comparative study. 
To learn realistic dialog policies, we need to predict the ASR confidence score that a dialog system would produce for a simulated ASR output. 
\citet{Schatzmann2007} and \citet{Thomson2012} have mentioned how to assign confidence scores to simulated ASR output, 
but their approaches simply sample from the training set distributions of ASR scores based on 
whether the simulated ASR output matches the reference without trying to predict the score based on the exact output. 
Also, to simulate errors on new data, we need to handle words outside the vocabulary of the 
n-gram confusion matrix constructed from training data. 
Hence, we improve upon the existing techniques by adding ASR confidence score prediction and out-of-vocabulary word mapping. 
Note that these improvements are not restricted to one particular technique like \citet{Schatzmann2007}. 
In fact, the ASR confidence score prediction from ASR output is also necessary for any phoneme-level error simulation.

Given transcribed training data with reference utterances, ASR hypotheses, and ASR confidence scores, 
we first construct the n-gram confusion matrix between ASR hypotheses and reference utterances 
after aligning each hypothesis with its reference by Levenshtein distance. 
To train the ASR score prediction model, 
we extract the Term-Frequency-Inverse-Document-Frequency (TFIDF) \cite{ManningCD2009} vectors for hypotheses and references separately, 
and compute WER related features including WER, sentence length, number of correct words, insertions, deletions, 
and substitutions for each hypothesis-reference pair. 
We then concatenate the TFIDF vectors and WER related features to form the input feature matrix. 
We can formulate the problem as either regression or classification on top of the feature matrix with the ground-truth ASR confidence scores as the targets.
One problem with regression is that the predictions cluster towards the mean score due to minimizing mean squared error, 
creating a clear difference between the distributions of predicted scores and actual scores. In the case of classification, 
we can first classify the score bin (e.g., bin $1$: [$0$, $0.1$), bin $2$: [$0.1$, $0.2$)), 
then sample from the training set distribution within the predicted bin to obtain the predicted score. 

On test/new reference utterances, we probabilistically partition each utterance into a list of n-grams 
based on their frequencies in the training data (see \cite{Schatzmann2007} for details), 
sample each n-gram's replacement (including itself which means no error) from the confusion matrix, 
and predict the ASR confidence score for each reference and simulated hypothesis pair with the trained ASR score model and the same feature extraction. 
Words not seen during training are mapped to their closest in-vocabulary counterparts through fuzzy-matching\footnote{\url{https://docs.python.org/3/library/difflib.html}}, 
but with some probability of staying the same to avoid inflating the substitution errors: Let the out-of-vocabulary word be $O$, its closest in-vocabulary match be $C$, and the WER in training data be $\widehat{P}$, then with probability $1 - \widehat{P}$ we let $O$ stay the same and otherwise sample its replacement from the confusion vector for $C$.

By design, the procedure produces WER and error type (insertion, deletion, substitution) distributions on test set similar to those of training set. 
To artificially increase or decrease WER, we can adjust the frequencies that each n-gram is substituted with itself (no error) in the confusion matrix.

\section{Experimental Results}

To compare different error simulation techniques, we used a corpus collected from MovieBot\footnote{\url{https://www.amazon.com/Amazon-MovieBot/dp/B01MRKGF5W}}, an Alexa Skill \cite{Kumar2017} that converses with users about existing, new, and forthcoming movies \cite{Fazel2017}.
We used a $75$/$25$ training/test split of the corpus which consists of $64,830$ transcribed user turns. 

\begin{table*}[!b]
  \centering
    \caption{Error distribution of real and simulated output on the test data}
  \begin{adjustbox}{max width=1.0\textwidth}
  \bgroup
  \def\arraystretch{1.0}
  \begin{tabularx}{0.86\linewidth}{lcccc}
    \toprule
    \multicolumn{1}{c}{} & \multicolumn{1}{c}{\textbf{Relative WER Change}} 
& \multicolumn{1}{c}{\textbf{Substitutions}} & \multicolumn{1}{c}{\textbf{Insertions}} & \multicolumn{1}{c}{\textbf{Deletions}} \\
    \midrule
ASR Output	&$ 0.00\%$& $48.32\%$&	$18.67\%$&	$33.01\%$ \\
	    \midrule 	
Audio-Level & $ +9.49\%$	&	$36.51\%$	&  $5.00\%$	&$58.49\%$ \\
	    \midrule
 Text-Level	&	$-10.40\%$ &$52.86\%$	& $16.29\%$ &$30.86\%$  \\
    \bottomrule
  \end{tabularx}
  \egroup
  \end{adjustbox}
  \label{moviebot-stats}
\end{table*}

\subsection{Analysis of Simulated ASR Hypotheses}

Table \ref{moviebot-stats} reports the relative WER change with respect to the real ASR output, as well as 
error type distributions for the real ASR output and the output of audio- and text-level error simulation for the test data.
In this table, we see that both methods match the WER of the real ASR output ($10\%$ relative change is small).
The distribution of errors in the text-level approach, however, better matches those of the real ASR output. 
For the audio-level method the errors are different between real and simulated outputs; 
whereas the majority of errors in real ASR are substitutions ($48\%$), the majority of errors in simulated ASR are deletions ($58\%$). 
This is partly due to the fact that WER in real data can be a result of factors other than background noise, such as accents; 
hence matching the WER in simulated data only through audio noise sometimes leads to too much noise and empty ASR output.

\begin{table}[t]
 \centering
    \caption{Evaluation of ASR confidence score prediction models}
 \begin{adjustbox}{max width=0.65\textwidth}
 \begin{tabular}{lcc}
 \toprule
         & \textbf{Linear Correlation} & \textbf{Mean Abs. Error} \\
 \midrule
	Baseline & $0.377$ & $0.241$ \\
	\midrule
	Regression & $0.777$ & $0.132$ \\
	\midrule
	Classification & $0.736$ & $0.150$ \\
 \bottomrule
 \end{tabular}
 \end{adjustbox}
  \label{table:score}
\end{table}

\begin{figure*}[!b]
  \centering
  \includegraphics[width=1.0\textwidth]{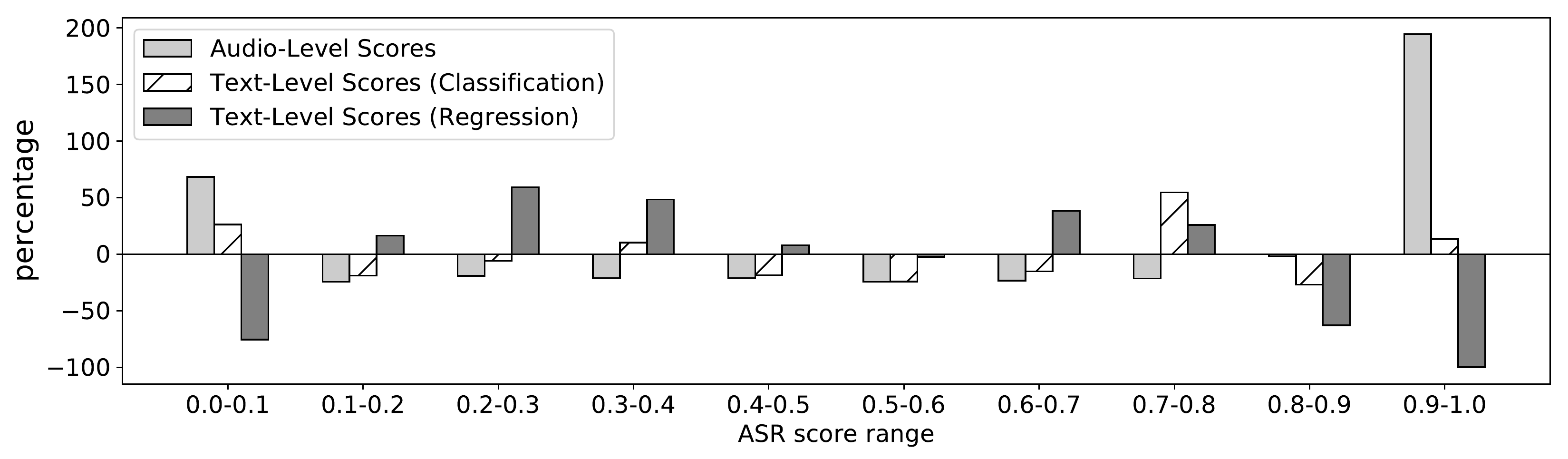}
  \caption{Relative changes in the simulated ASR scores compared to the real scores on the test set}
  \label{fig:score-dist}
\end{figure*}

\subsection{ASR Confidence Scores for Simulated ASR Hypotheses}
\label{subsec:score-eval}

We experimented with both classification and regression models for ASR score prediction for text-level error simulation, 
and evaluated it by comparing its predictions on the real ASR hypotheses with the ground truth ASR scores in the test set.
We used \citeauthor{Schatzmann2007}'s \cite{Schatzmann2007} approach as baseline, 
which samples from the training set score distribution based on whether the hypothesis matches the reference.
After preliminary experimentation we chose gradient boosting \cite{FriedmanJ2001} for both classification and regression. 
It is possible to explore deep learning models for score prediction, but for relatively simple tasks, 
TFIDF with an off-the-shelf classifier can produce competitive results \cite{TongE2017, WangL2018}.  

Table \ref{table:score} shows the linear correlation and Mean Absolute Errors against the real ASR scores for each model.
Both regression and classification approaches substantially outperformed the baseline and yielded predictions 
that are close to the real scores in terms of correlation and MAE. 
The predictions from the classification approach match the ground-truth scores much better in distribution 
but have slightly smaller correlation and larger MAE compared to the predictions from the regression approach.
It is worth pointing out that regardless of the score model, there is an inherent gap 
between scores predicted from the hypothesis-reference pairs and real scores, 
because for the same hypothesis-reference pair, the real score could still vary based on the audio quality. 

Figure \ref{fig:score-dist} illustrates the relative changes in the simulated ASR scores compared to the real scores on the test set. 
We see that the text-level scores with regression are skewed towards the center as a result of minimizing mean squared error. 
The distributions for the audio- and text-level scores with classification are both close to that of the real scores except for rare spikes, 
with a Kullback-Leibler divergence of $0.08$ and $0.03$, respectively. 
For the audio-based method, the spike in the $0.0$-$0.1$ bin is due to cases where too much noise is added and as a result the ASR model considers the input as noise; 
the spike in the $0.9$-$1.0$ bin, on the other hand, is due to the characteristics of the TTS output where in certain cases the words are clearly spoken and the added noise does not distort the signal.
For text-level scores with classification, the spikes can be explained by the model always predicting the most frequently seen bin for hypothesis-reference pairs that occur multiple times.
When we examined the samples that are misclassified to the $0.7$-$0.8$ and $0.0$-$0.1$ bins in the training set, 
we found that $96\%$ are hypothesis-reference pairs with multiple duplicates that are most frequently in the respective bins 
but also spread across other score bins.

\subsection{Discriminator}
\label{discriminator-section}

\begin{table*}[t]
  \centering
   \caption{Discriminator performance (poorer performance means more realistically simulated output)}
  \begin{adjustbox}{max width=1.0\textwidth}
  \bgroup
  \def\arraystretch{1}
  \begin{tabularx}{1.05\linewidth}{lcccc}
    \toprule
          & \textbf{Accuracy} & \textbf{Precision} & \textbf{Recall} & \textbf{F-Score} \\
         \midrule
	Audio-Level, Without Score & $62.59\%$ & $0.688$ & $0.462$ & $0.552$ \\
	Text-Level, Without Score & $60.36\%$ & $0.633$ & $0.492$ & $0.554$ \\
	\midrule
	Audio-Level, With Score & $69.01\%$ & $0.725$ & $0.614$ & $0.665$ \\
	Text-Level, With Score (Classification) & $75.18\%$ & $0.727$ & $0.807$ & $0.765$ \\
	Text-Level, With Score (Regression) & $88.30\%$ & $0.855$ & $0.923$ & $0.888$ \\
	\midrule
	Audio-Level, Without Score, No Duplicates & $63.03\%$ & $0.619$ & $0.679$ & $0.647$ \\
	Text-Level, Without Score, No Duplicates & $64.95\%$ & $0.642$ & $0.676$ & $0.659$ \\
	\midrule
	Audio-Level, With Score, No Duplicates & $65.54\%$ & $0.665$ & $0.628$ & $0.646$ \\
	Text-Level, With Score (Classification), No Duplicates & $67.05\%$ & $0.665$ & $0.689$ & $0.676$ \\
	Text-Level, With Score (Regression), No Duplicates & $78.33\%$ & $0.752$ & $0.846$ & $0.796$ \\
    \bottomrule
  \end{tabularx}
  \egroup
  \end{adjustbox}
  \label{table:discriminator}
\end{table*}

To evaluate how realistic the simulated ASR output is for each error simulation method, 
we adopted \citeauthor{Wang2018}'s \cite{Wang2018} approach of training a discriminator to see 
how well it can distinguish the simulated ASR output from the real one. 
We are interested in seeing how realistic the simulated ASR hypotheses are, 
both on their own and together with the (predicted) ASR scores, 
so we ran experiments with and without ASR scores in the input. 
\citeauthor{Wang2018}'s \cite{Wang2018} discriminator takes the ASR hypotheses as input, 
but we postulate that to better distinguish whether a hypothesis is simulated, 
the discriminator should compare it with the reference 
(which was confirmed in our preliminary experiments where adding the reference improved the accuracy by \char`~$2\%$ for the cases considered), 
hence we included references in the input.
We extracted TFIDF vectors from both hypotheses and references, computed WER related features, 
and optionally concatenated them with ASR scores to form the input matrix, 
then we fitted a Gradient Boosting Classification model on top to predict whether each hypothesis is simulated or not.

Table \ref{table:discriminator} shows the test set performance of the discriminator in terms of accuracy, precision, recall, 
and F-score for different combinations of error simulation methods, input features, and data preprocessing. 
Poorer discriminator performance means more realistically simulated ASR output. 
Disregarding ASR scores, the discriminator has low accuracy on both text- and audio-level methods at $60\%$ and $63\%$, respectively, 
indicating that the simulated hypotheses from both methods are realistic. 
Including ASR scores as a predictor, the accuracy for audio-level method improves to $69\%$ despite the fact that 
its ASR scores are obtained through ASR decoding, 
but the increase can be likely explained by the score distribution difference seen in Figure \ref{fig:score-dist}. 
In comparison, the accuracy for both text-level methods have much larger increases.
For text-level method with regression, it is easy to see why the scores can be a strong predictor, 
given the large score distribution differences between the real and predicted scores by this method (Figure \ref{fig:score-dist}). 
For text-level method with classification,
the reason is more subtle and a result of the score model only predicting the most frequent bin for hypothesis-reference pairs that have multiple duplicates with different real ASR scores. 
For a more details see Appendix \ref{appendix-classification-example}.

To evaluate how realistic the score predictions are without the superficial influence of duplicate hypothesis-reference pairs, 
we repeated the experiment after removing such pairs in real and simulated data (last five rows in Table \ref{table:discriminator}). 
The training and test set sizes reduced to $20,721$ and $7,639$, respectively. 
We see that the discriminator performance for the text-level method with classification is now very close to when scores are excluded.

\subsection{Effect on Natural Language Understanding}

\begin{table*}[t]
  \centering
            \caption{Relative error rate change for in- and out-of-domain requests with respect to the reference.}
  \begin{adjustbox}{max width=1.0\textwidth}
  \bgroup
  \def\arraystretch{1.0}
  \begin{tabularx}{1.2\linewidth}{lcccc}
    \toprule
        \multicolumn{1}{c}{} & \multicolumn{1}{c}{\textbf{Relative Semantic}}  & \multicolumn{1}{c}{\textbf{Relative Intent}} & \multicolumn{1}{c}{\textbf{Relative Slot}} & \multicolumn{1}{c}{\textbf{Relative Out of Domain}}  \\
    \multicolumn{1}{c}{} & \multicolumn{1}{c}{\textbf{Error Rate Change}}  & \multicolumn{1}{c}{\textbf{Error Rate Change}} & \multicolumn{1}{c}{\textbf{Error Rate Change}} & \multicolumn{1}{c}{\textbf{Error Rate Change}}  \\
    \midrule
	Real ASR Output & $46.51\%$ & $58.24\%$ & $62.62\%$ & $2.65\%$\\
	\midrule
	Audio-Level & $71.61\%$ & $109.90\%$ & $76.78\%$ & $1.89\%$\\
	\midrule
	Text-Level & $65.72\%$ & $92.75\%$ &$62.43\%$ &$2.89\%$\\
    \bottomrule
  \end{tabularx}
  \egroup
  \end{adjustbox}
  \label{nlu-stats}
\end{table*}

To understand the effect of each method on the NLU output, we computed the relative semantic error rate (SER), intent error rate, and slot error rate with respect to the reference data by calculating the exact intent and slot matches for 
in-domain and out-of-domain\footnote{$23\%$ of the annotated data are out-of-domain requests.}  requests (Table \ref{nlu-stats}). 
This analysis was done on the $43\%$ of the test data that was annotated for intent and slots. 
For the audio-level method, the higher relative SER change can be attributed to the higher number of word deletions compared to the the real ASR output.  
For the text-level method, $15\%$ of the errors are related to one-word utterances. 
When an n-gram gets confused with a single word, it can create unrealistic entries in the confusion matrix for its subset words. 
Pruning the confusion matrix with phone level similarity, Levenshtein distance, or minimum frequency can reduce the unrealistic entries, 
but the remaining frequencies need to be re-weighted accordingly after pruning to maintain the same WER.

\subsection{Comparison of Learned Dialog Policies}

In this section we report preliminary results for learning dialog policies with a user simulator and each error simulation technique. 
We used a user simulator based on the approach presented in \citep{Fazel2017} and trained error recovery policies using deep RL.
More specifically, the components used to learn the policies are as follows. 
The state is composed of 1) hypothesis intent and slot, 2) ASR confidence score, 3) previous agent action, 4) total number of clarifications so far in the dialog, and 5) number of clarifying questions for the current user request, where hypothesis is the agent's recognition. 
The action space is composed of three actions that are critical for conversational error recovery: \textit{execute}, \textit{confirm}, and \textit{repeat}. 
\textit{Execute} performs a backend action (e.g., accesses a database) and answers the user's question. 
Clarifying questions (\textit{confirm},  \textit{repeat}) are used to recover from errors. 
For the agent's policy, we used Dueling DDQN \cite{WangZiyu2016} with a fully-connected MLP to represent the deep Q-network. 
The hidden layers use a rectifier nonlinearity, and the output layer is a fully connected layer with linear activation function and a single output for each valid action. 
Finally, the reward function is summarized in Table \ref{reward_function}. 
Here, \textit{reference} indicates the actual intent and slot as generated by the user simulator, and \textit{hypothesis} refers to what the agent understands.
If \textit{hypothesis} $\neq$ \textit{reference}, this indicates that ASR/NLU errors were induced. 
The agents gets a positive reward for executing correctly, a negative reward for executing incorrectly, and smaller negative rewards for confirm and repeat. 
A successful execution is when the reference and hypothesis intent and slots are equal. 
The penalty for \textit{repeat} is higher because it shows that the agent didn't understand any of what the user said. 
Additionally, if the user says an utterance with positive sentiment, the agent receives a small positive reward, 
but if the user interrupts the agent (a barge-in) or responds with an utterance with negative sentiment, the agent receives a small negative reward. 
Note that the state-based rewards are added to the action-based rewards. 
For more details see Appendix \ref{ddqn}.

\begin{table}[!b]
  \centering
          \caption{Reward Function}
  \begin{adjustbox}{max width=0.70\textwidth}
  \begin{tabular}{l|c|c}
    \toprule
    \multicolumn{1}{c|}{\textit{a}}  & \textit{s} & \textit{r} \\
    \midrule
	execute(\textit{hypothesis}) & $hypothesis = reference$ & $+1$ \\
	execute(\textit{hypothesis}) & $hypothesis \neq reference$ & $-1$ \\
	confirm & $*$ & $-0.33$ \\
	repeat & $*$ & $-0.50$ \\
	\midrule
	\multicolumn{1}{c|}{$*$} & positive sentiment & $+0.17$ \\
	\multicolumn{1}{c|}{$*$} & negative sentiment & $-0.17$ \\
	\multicolumn{1}{c|}{$*$}& barge-in & $-0.17$ \\
    \bottomrule
  \end{tabular}
  \end{adjustbox}
      \label{reward_function}
\end{table}

\subsubsection{Simulation Results}

\begin{figure}[t]
  \centering
    	\begin{subfigure}{1.0\linewidth}
    	\centering
  		 \includegraphics[width=0.7\textwidth]{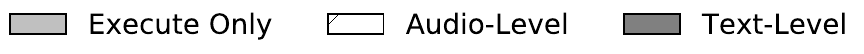}
    \end{subfigure} 
    \hspace{1mm}  
  	\begin{subfigure}{0.3\linewidth}
  		\includegraphics[width=0.6\textwidth]{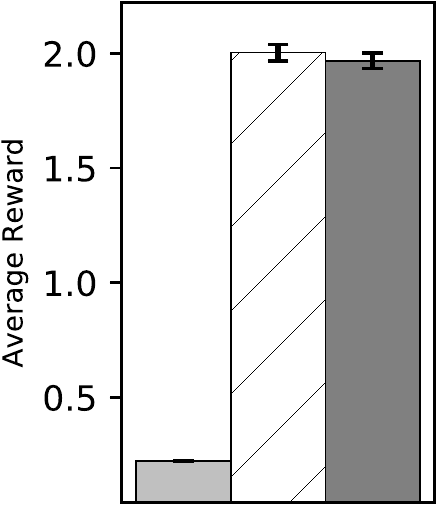}
  	\caption{}
  	\label{average_reward_test}
    \end{subfigure}
    \hspace{1mm}
  	\begin{subfigure}{0.3\linewidth}
  		\includegraphics[width=0.6\textwidth]{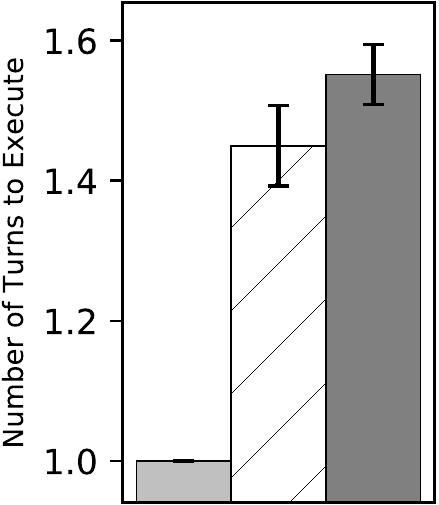}
  	\caption{}
  	\label{average_turn_test}
    \end{subfigure}
    \hspace{1mm}
  	\begin{subfigure}{0.31\linewidth}
  		\includegraphics[width=0.6\textwidth]{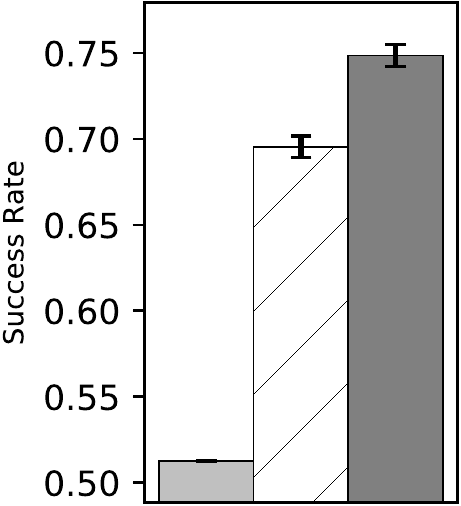}
  	\caption{}
  	\label{average_success_test}
    \end{subfigure}
  \caption{Test performance of the dialog policies averaged over $2,400$ dialogs. a) average reward, b) number of turns to execute, and c) success rate.}
  \label{comparison}
\end{figure}

To compare the policies learned with the audio- and text-level techniques, we used the user simulator with ASR output from real users and tested each agent on $2,400$ dialogs.
To achieve a fair comparison, we fixed the random number generator so that the user simulator generated identical sequences of intents and utterances for both agents.
As a baseline, we evaluated these dialogs with a policy that only executes. 
Figure \ref{comparison} shows the results. 
The Y-axis in the three subfigures is the average reward, the average number of turns to execute, and the execution success rate of each technique, respectively. 
Average number of turns to execute is the number of clarification turns needed to execute for each agent, where $1$ means execute without asking any clarifying question. 
Overall, policies learned using both audio- and text-level error simulators achieved significantly higher success rates compared to the execute-only policy, showing the importance of using a clarification capability.
The  text-level policy achieved a higher success rate, and was able to identify more incorrect turns ($51\%$) compared to the audio-based policy ($41\%$), however, this came at the cost of asking more clarifying questions.

\subsubsection{User Study}

\begin{table}[b!]
  \centering
          \caption{Cell and marginal means of the dialog ratings}
  \begin{adjustbox}{max width=1.0\columnwidth}
  \begin{tabular}{lcc|c}
    \toprule
     & \textbf{Audio-level} & \textbf{Text-level} & \\
    \midrule
    Volume Low & $3.06$ & $3.38$ & $3.22$ \\
    \midrule
    Volume High & $3.13$ & $2.81$ & $2.97$ \\
    \midrule
     & $3.09$ & $3.09$ & $3.09$ \\
    \bottomrule
  \end{tabular}
  \end{adjustbox}

      \label{dialog_ratings}
\end{table}

We also conducted a user study to compare the real-world performances of the dialog policies 
learned with audio- and text-level techniques. 
To increase the statistical power given the limited number of subjects, 
we adopted the within-subjects experimental design and analysis \cite{LaneD2013}, 
where each subject was assigned with and measured on all possible treatment combinations. 
More specifically, we recruited $16$ internal subjects and 
considered two treatment factors: noise volume (high, low) and error simulation technique (audio-level, text-level), 
for a total of four unique treatment combinations. 
The sample size and number of treatment combinations are 
comparable to previously published studies under the same experimental design \cite{PillaR2005, FestingM2014}. 
We asked each subject to engage in four dialogs with MovieBot, one for each treatment combination. 
For example, if the treatment combination was "high noise volume + text-level technique", 
the subject would talk to the SLU application that adopted a dialog policy learned with the text-level error simulation technique, 
while a sound track of people talking on high volume was played in the background.
We asked subjects to rate the quality of their most recent dialog on a $5$-point scale at the end of each dialog. 
We followed the standard practice for conducting within-subjects experiments, 
including counter-balancing the order of treatment combinations.
The cell and marginal means of the dialog ratings collected are shown in Table \ref{dialog_ratings}. 
Note that by design the observations from the same subject are correlated. 
We performed the two-way repeated measures ANOVA \cite{LaneD2013}. 
The partial output is shown in Appendix \ref{ANOVA}. 
The audio- and text-level techniques have the same marginal mean ratings ($3.09$), 
hence there is $0$ main effect for "technique" in the ANOVA. 
The audio-level technique has lower mean ratings than text-level for low noise volume, 
and vice versa for high noise volume. 
However, the differences are not statistically significant based on the 
\textit{p}-value of the interaction effect of "noise $\times$ technique" in the ANOVA. 
Overall, the quantitative results from the user study do not reveal significant differences between the two techniques.

\section{Conclusion}

In this paper, we compared different error simulation techniques, with a focus on audio- and text-level methods,
to assess their suitability for learning clarification dialog policies to recover from errors.
In a variety of metrics considered, both the audio- and text-level methods produced similar results
which are reasonably close to the ground truth.
Therefore, the text-level method should be the preferred method for policy learning 
due to its simplicity and cheap computational cost.

Future work should focus on identification of types of errors that are recoverable, 
better detection of out-of-domain and out-of-vocabulary requests, 
and investigation of correlations between different sources of noise and the domain of user's requests.
We plan to improve the existing audio-level method by inserting different levels and types of noise 
based on the audio content instead of uniform-randomly, such as inserting music noise when the user asks to change the song.
Additionally, we plan to further improve the existing text-level method by leveraging phoneme/semantic similarity 
when mapping out-of-vocabulary words or trimming the n-gram confusion matrix. 
We can also explore generative adversarial networks \cite{GoodfellowI2013} for simulating more realistic ASR output.

\subsubsection*{Acknowledgments}

We would like to thank Shang-wen Li, Jin Cao and Jared Casale for helping set up the audio-based error simulation. We would also like to thank John Peach, Jason Pazis, Josep Valls Vargas, Lazaros Polymenakos, and Praveen Bodigutla for their feedback and support. 

\section*{Appendix}
\renewcommand{\thesubsection}{\Alph{subsection}}

\subsection{\citeauthor{Schatzmann2007}'s (\citeyear{Schatzmann2007}) N-Gram Confusion Matrix Approach}
\label{schatzmann-approach}
We briefly describe the text-level error simulation approach by Schatzmann et al. (\citeyear{Schatzmann2007}),
which served as the starting point for our approach described in Section \ref{subsec:text-sim}:
\begin{itemize}
\item During training: For each reference utterance and ASR hypothesis pair from the training data, 
align the reference fragments (a ``fragment'' is essentially an n-gram for an arbitrary n) 
with hypothesis fragments by minimizing the Levenshtein distance. 
For each pair of aligned fragments, increment their frequency in the confusion matrix. 
Also, count the frequencies of all fragments in the training data.
\item During testing: For each reference utterance in the test/new data, 
partition the sentence into fragments sequentially and probabilistically 
based on their frequencies in the training data: 
Moving from the first word to the last word in the utterance, 
a word belongs to the previous fragment with probability 
$freq$(previous fragment joined with the current word) / $freq$(previous fragment), 
otherwise it starts the next fragment. 
After partitioning, sample a replacement (including the original fragment in case of no error) 
for each fragment from the confusion matrix to generate the ASR hypothesis. 
Note that Schatzmann et al. did not mention how to handle out-of-vocabulary fragments. 
Lastly, sample an ASR score for the simulated ASR hypothesis from the training data ASR score distributions 
based on whether the simulated ASR hypothesis contains any word errors.
\end{itemize}

\subsection{Effect of Score on the Discriminator}
\label{appendix-classification-example}

In Section \ref{discriminator-section}, we saw that (predicted) scores improve the accuracy of the discriminator. 
For text-level method with classification the reason is more subtle.
Recall that in this approach we sample from the training set distribution within the classified score bin so there is no loss in granularity.
As mentioned in Section \ref{subsec:score-eval}, a certain hypothesis-reference pair $U$ can have multiple duplicates with different real ASR scores, 
but the score model only predicts the most frequent bin. 
Now, let's say in the training set $U$ appears $100$ times in both real and simulated data, 
its real ASR scores are most frequently in bin 8, which happens $30\%$ of the time. 
In this case, the score model always predicts bin 8 for $U$. 
The discriminator learned the following rule - "if the score of $U$ is in bin 8, $U$ is simulated, and otherwise not simulated". 
The discriminator would then be able to correctly predict $100\%$ of $U$ in the simulated data and $70\%$ of $U$ in the real data, 
and achieve $(70 + 100) / 200 = 85\%$ training set accuracy. 
Because the test set is similar to the training set, 
and we know that there are many hypothesis-reference pairs similar to $U$ from the investigation of the spikes in Figure \ref{fig:score-dist}, 
it follows that the discriminator can have high test accuracy using the ASR scores as a predictor.

\subsection{Details of Policy Learning}
\label{ddqn}

We used Dueling DDQN \cite{WangZiyu2016} with a fully-connected MLP to represent the deep Q-network. 
The hidden layers use a rectifier nonlinearity, and the output layer is a fully connected layer with linear activation function and a single output for each valid action
We trained the agent using an $\epsilon$-greedy policy with $\epsilon$ decreasing linearly from $1$ to $0.1$ over $100,000$ steps. 
Additionally, we tuned a window size to include previous dialog turns as input and set $\gamma$ to $0.97$. 
We ran the method $30$ times for $150,000$ steps, and in each run, after every 10,000 steps, 
we sampled $100$ dialog episodes with no exploration to evaluate the performance.
The optimal parameters were found using Hyperopt \cite{hyperopt}.
The learning parameters for each error simulator are show in the following table.

\begin{table}[H]
\small
  \centering
  \label{parameters}
        \begin{adjustbox}{max width=\textwidth}
  \begin{tabular}{l|c|c}
    \toprule
    \multicolumn{1}{c|}{}  & \multicolumn{1}{c|}{Audio-Level} & \multicolumn{1}{c}{Text-Level} \\
    \midrule
	Number of Hidden Layers & $2$ & $2$ \\
	Number of Hidden Nodes & $64$ & $128$  \\
	Learning Rate & $0.0001$ & $0.0001$  \\	
	Dropout & $0.1$ & $0.5$ \\
	Experience Replay Size & $10,000$ & $15,000$ \\
	Embedding Size & $1$ & $20$  \\
	Target Model Update Interval & $13,000$ &$9,000$  \\
    \bottomrule
  \end{tabular}
  \end{adjustbox}
\end{table}

\subsection{Two-Way Repeated Measures ANOVA on the Dialog Ratings in the User Study}
\label{ANOVA}

\begin{table}[h]
  \centering
  \begin{adjustbox}{max width=0.8\textwidth}
  \begin{tabular}{lcccc}
    \toprule
    \textbf{Source} & \textbf{Sum} & \textbf{Degrees} & \textbf{F Value} & \textbf{p-value} \\
    & \textbf{of} & \textbf{of} &  &  \\
    & \textbf{Squares} & \textbf{Freedom} &  &  \\
    \midrule
    Noise Volume & $1$ & $1$ & $1.071$ & $0.317$ \\
    \midrule
    Error Simulation Technique & $0$ & $1$ & $0$ & $1$ \\
    \midrule
    Noise $\times$ Technique & $1.562$ & $1$ & $3.641$ & $0.076$ \\
    \bottomrule
  \end{tabular}
  \end{adjustbox}
\end{table}

\bibliographystyle{named}
\bibliography{neurips_2019}

\end{document}